\documentclass[letterpaper, 10 pt, conference]{ieeeconf}  

\usepackage{microtype}
\usepackage{graphicx}
\usepackage{subfigure}
\usepackage{cite}
\usepackage{booktabs} 
\usepackage{amsmath,amssymb,amsfonts}
\usepackage{booktabs}
\usepackage{fixmath}
\usepackage{svg}
\usepackage{textcomp}
\usepackage{gensymb}
\usepackage{textcomp}
\usepackage{mathtools}
\usepackage[export]{adjustbox}
\usepackage{float}
\usepackage{multirow}
\usepackage{tabularray}
\usepackage[hidelinks]{hyperref}
\usepackage[flushleft]{threeparttable}

\usepackage{wrapfig}
\usepackage{pifont}
\usepackage{bm}
\usepackage{balance}
\usepackage[flushleft]{threeparttable} %

\IEEEoverridecommandlockouts                              

\def\projname{MiBOT}

\title{\LARGE \bf
MiBOT: A head-worn robot that modulates cardiovascular responses through human-like soft massage
}

\author{%
Alice Mylaeus$^{1,*}$,
Stephanie Vogt$^{2,*}$,
Berken Utku Demirel$^{1}$,
Marcel Gort$^{2}$, \\
Mirko Meboldt$^{2}$,
Manuel Meier$^{1}$, and
Christian Holz$^{1}$
\thanks{* These authors contributed equally to this work.}
\thanks{%
$^{1}$Department of Computer Science, ETH Zürich, Switzerland.
$^{2}$Department of Mechanical Engineering, ETH Zürich, Switzerland.
This work was not supported by any organization.
}
}

\begin{document}

\maketitle
\thispagestyle{empty}
\pagestyle{empty}

\begin{abstract}
Massage therapy is helpful for the rehabilitation of various diseases, such as headaches caused by migraines and stress. 
Existing robotic systems have focused on massage therapy on the torso and limbs, but performing massage motions through suitable actuation on a person's head has been a challenge.
In this paper, we present \emph{MiBOT}, a head-worn massage robot that actuates two soft tactors to produce touch motions mimicking human massage.
A key design principle behind MiBOT is its silent actuation, which we achieve through pneumatic artificial muscles in conjunction with a controller loop to respond to contact pressure. 
We evaluated the effectiveness of MiBOT in a controlled study and assessed subjects' blood pressure and heart rate levels while applying MiBOT.
We found that our mechanical system generated positive and conclusive quantitative outcomes that are similar to the human-administered massage, decreasing participants' mean systolic and diastolic blood pressure by 2.8\,mmHg and 1.7\,mmHg, respectively, as well as calming their heart rate by 8--10\% on average. 
\end{abstract}

\section{Introduction}

According to the Global Burden of Disease (GBD) study, headache disorders, which include migraine and tension-type headaches, are among the most prevalent disorders of mankind and disabling conditions worldwide~\cite{stovner_global_2022}. 
The World Health Organization (WHO) estimates that the prevalence among adults of current headache disorder (symptomatic at least once within the last year) is about 50\%. 
Studies have estimated that half to three-quarters of adults aged 18–-65 years in the world have had headaches in the last years and, among those individuals, 30\% or more have reported migraine. 
Headache on 15 or more days every month affects 1.7--4\% of the world’s adult population~\cite{ATLAS}.

\begin{figure}[t]
    \centering
    \includegraphics{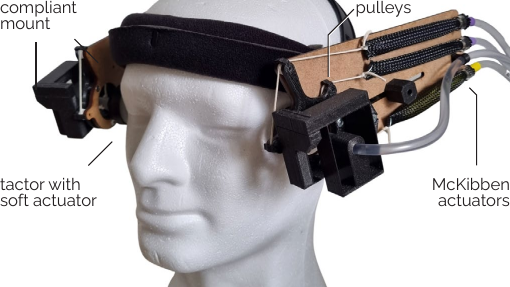}
    \caption{\emph{\projname} is a head-massage robot that integrates two symmetrically actuated massaging units, each mounting pneumatic artificial McKibben muscles to actuate a soft tactor on either side of the face.
    \projname\ dynamically adjusts the pressure applied to the wearer's head through a sensor-driven control loop to dynamically conform to various head shapes and topological features, while performing massage motions reminiscent of human touch.}
    \label{fig:end_device}
\end{figure}

Some studies found that massage may be as effective as medication in the preventative management of headaches~\cite{migraine_headache}.
This is particularly true of those caused by migraines, where the activation of the trigeminal nerve releases calcitonin gene-related peptide (CGRP) from the trigeminovascular system~\cite{EDVINSSON2012319} and causes the release of pain-producing inflammatory substances around the nerves and blood vessels of the head~\cite{brain_int}. 
Massage therapy is a manipulation of the soft tissue in humans by causing movement or applying pressure to the body.
It improves blood circulation within the body, thus promoting healing in the patient~\cite{old_massage, ICRA_peristaltic, ICRA_emg}.
Moreover, chronic excessive use of medication to treat headaches can cause even more frequent headache episodes which, again, increases the need for medication, leading to persistent headaches in the end-stage~\cite{ATLAS}.
This makes massage therapy the preferable solution to break this cycle.
Massages are also an important part of healthy lifestyles to relieve stress and relax the body~\cite{massage_frontiers, ICRA_HRV_back} while reducing chronic pain, anxiety, and depression and improving the quality of sleep and life~\cite{IROS_EEG_massage,ICRA_lately}.

At present, most types of massage actions require massage therapists. 
However, such manual massages require considerable resources, most notably labor cost and time~\cite{ICRA_contact, hair_wash, Iros_new}.
Furthermore, it is challenging for therapists to continuously exert suitable levels of force for a long time without fatigue.
This decreases the number of treated patients and the effect of therapies.
Thus, developing massage robots with human-like touch has become an important research topic in robotics to produce automated, repeatable massages that could become a standard part of therapy~\cite{old_massage, massage_frontiers, Dong_IROS, ICRA_soft}.
Here, a self-actuated robotic system would be capable of operating without fatigue and have the potential to consistently treat a large number of  patients~\cite{old_massage}.

To date, several studies designed a mechanical therapy unit and verified the feasibility of developing an intelligent massage robot~\cite{Kume}. 
For example, Jones et al.'s developed robotic system focused on the lumbar and thoracic portions of a patients' back, generating proper trajectories and controlling motions such that the robot produced optimal massage movements with proper force~\cite{old_massage}.
Luo et al. leveraged electromyographic (EMG) signals as part of a multi-finger robot hand control to create human-like massage on a person's shoulder~\cite{EMG_massage}, analyzing EMG signals to distinguish painful and comfortable feelings.
Depending on a person's feelings and exerted massage force, their system obtains the appropriate range of input for the robotic hand. 
Peng et al. imitated human palm and finger motions to produce actions of massage including ``pushing, picking up, and kneading''~\cite{ECG_massage}.
Luo et al. proposed an anthropomorphic dual-arm robot that utilizes techniques through the impedance control of both cartesian space and joint space~\cite{IROS_EEG_massage}.
They evaluated massages using participants' electroencephalogram (EEG) signals during therapy.

Previous studies on human-like massage have focused on the back, shoulders, and arms (e.g.,~\cite{old_massage, IROS_EEG_massage, EMG_massage, ECG_massage, IROS_Normal_Massage, ICRA_Arm, ICRA_back, IROS_OLD}).
They may not apply to headaches or specifically migraine and tension-related headaches, as they are caused by inflammatory substances around the head's blood vessels~\cite{ATLAS}. 
Previous massage robots that targeted the temple region have not been evaluated for the robot's effect on the person's cardiovascular system to measure the sympathetic response and, thus, the quantified outcome.
Since massage therapy is effective in reducing blood pressure (BP) and heart rate (HR)~\cite{elsevier_bp_hr}, we argue that, optimally, a massage robot would produce similar effects compared with human-administered massage.
In other words, for a massage robot to be considered successful and beneficial, it should be able to reduce blood pressure and heart rate to a comparable extent as a human massage.

\begin{figure}
    \centering
    \includegraphics{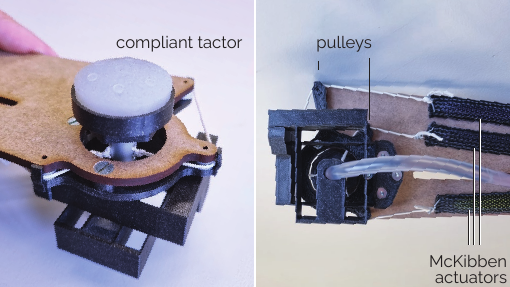}
    \caption{The compliant tactor and the pulleys for providing guidance and force transfer.
    \projname\ integrates compliant mechanisms~(a) and pulleys~(b) to transform unidirectional motion into circular massage motions that approximates human-like touch.}
    \label{fig:actuation}
\end{figure}

In this paper, we introduce \emph{\projname}, a novel wearable massage robot that produces human-like massage sensations on a person's temple regions.
\projname, where the overall design is shown in Fig.~\ref{fig:end_device}, operates at a non-disturbing noise level and adapts to complex target topologies as well as a wide range of different head shapes.
We evaluate the effect of the presented robot system on nine participants by monitoring their heart rate (HR) and blood pressure values while comparing the same quantitative metrics with a commercially available head massage.
Overall, our results indicate that the massage delivered by the designed robot can modulate the cardiovascular system while showing a significant relaxation effect on participants.

We make the following contributions in this paper:

\begin{itemize}
    \item \projname, a novel human-like massage robot system that administers soft tactor actuation on the temple region of participants to alleviate headaches,

    \item a feedback control loop scheme that allows massage motions to follow a defined trajectory with proper force despite naturally varying head topologies, and

    \item an experimental evaluation of \projname's efficacy and effect on cardiovascular responses with nine participants. 
    Our results show that our robotic system decreased participants' heart rates and blood pressure levels. 
    To the best of our knowledge, we are the first to evaluate the effect of a wearable massage robot targeting the temple region on the human cardiovascular system and \emph{quantitatively} demonstrate its effect.
\end{itemize}


\section{Materials and Methods}
\label{sec:Prop_method}

\begin{figure}[t]
    \centering
    \includegraphics{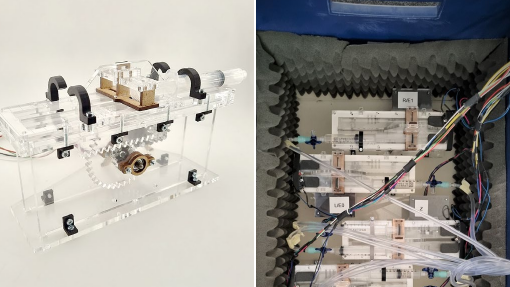}
    \caption{(left)~An actuation block with the stepper motor, 3:1 gear reduction, and a rack to compress the mounted syringes. 
    (right)~A sound-insulated box houses five actuation blocks; three drive two syringes for circular massage motion and two drive one syringe each for tactor pressure adaptation.}
    \label{fig:box_motor}
\end{figure}

\subsection{Design Concept and Hardware Setup}

Our massage robot is composed of 4 main parts. 
The first part is an actuation block that drives the whole system with the massaging motion without the need for human input.
The second is an actuation transmission block that allows the physical separation of the actuation unit and massaging locations that are in close proximity to the patient’s ear.
This is important to reach an acceptable noise level.
Since our novel massage robot focuses on a head area different from previous works, the reduction of vibrations and noise level is crucial to prevent users from experiencing physical discomfort.
The third main part is an actuation output part that,  depending on the chosen approach, either produces the massaging motion directly or some other form of motion that can later be transformed into a massaging motion.
Lastly, the system is expanded by guidance and force transfer block that allows the aforementioned transformation, e.g., from linear motion to circular massaging motion.
We explain each part of the designed system in detail as follows.

\subsubsection{Actuation Unit}
For the actuation unit, we use electrical stepper motors (SY42STH47, JOY-IT,  31.38\,N.cm holding torque) for their high torque at low speeds, feasibility, and availability.
Although a linear magnetic actuator can be virtually silent~\cite{magnetic_actuators}, magnetic actuators have a high cost compared to electrical motors.
Therefore, we embed stepper motors into an adjunct custom-made acrylic transmission system pictured in Fig.~\ref{fig:box_motor}, and gear reduction with rotary to linear motion transformation is enabled.
This in turn allows axial compression of mounted standard syringes that act as fluidic cylinders and are connected to the massaging device in a closed system.

\subsubsection{Actuation Transmission}
Since the reduction of noise and vibrations are two important factors in making our system comfortable to the user, directly attaching the actuation unit to the headpiece of the massaging device is not feasible.
We, therefore, investigated pneumatic and hydraulic transmission systems and observed that pneumatic systems have a significant drawback which is the aforementioned high noise emissions.
Thus, we use a hydraulic transmission in our final system.
In our final robotic system, we measured the noise of actuation in a quiet environment (no external sound source is available in the environment during the measurement) with a sound level meter at participants' head height resulting in a maximal noise level of 38\,dB which is comparable to a soft whisper~\cite{Noise_Level}, proving our designed actuation transmission component reduces the overall noise emission.

\subsubsection{Actuation Output}
We employ McKibben pneumatic artificial muscles to fulfill the function of producing the actuation output since it is suitable for use in an environment with people as the driven part of this actuator is flexible~\cite{Mckibben}.
In Equation~\ref{eq:Mc}, the formulated static model characterizes the tension force $F$ of a McKibben Actuator in relation to the applied pressure $P$ and the knitting angle $\Theta$.
\begin{equation}\label{eq:Mc}
    F = \frac{\pi D_0^2 P}{4}(3\cos^2\Theta-1),
\end{equation}
where $D_0$ is the diameter when $\Theta$ equals 90$^{\circ}$. 
The tension force can be regarded as linearly proportional to the applied internal pressure and as a monotonic function of the knitting angle.
Maximal shortening can be calculated by solving the equation for $F = 0$, meaning the exerted tension force reaches zero. Hence, this state is reached when the knitting angle reaches $\Theta = 54.7 ^{\circ}$, also called the neutral angle. 
The prototyped McKibben actuators of \projname\ exhibit resting knitting angles of $28 ^{\circ}$, where its force $F_m$ under a certain pressure $P$ is modeled as:

\begin{equation}
F_m(P) = P \times A \times \tan(\Theta),
\end{equation}
where $A$ is the cross-sectional area of McKibben actuators.

Since an individual McKibben actuator generates muscle-like contractile motions~\cite{McKibben_Japan}, it can only provide unidirectional tension forces. 
Therefore, this sort of actuation output needs to be combined with further guidance and force transfer elements in order to be able to generate the aspired massaging motion and forces to the temple region of the head.

\subsubsection{Guidance and Force Transfer}
We use pulleys and compliant mechanisms in the last function block to transform unidirectional motion, e.g., provided by several McKibben actuators, into a circular massaging motion while providing the necessary limitation to the range of motion.

The overall designed massage robot with the two symmetrical sides of the massaging mechanism is comprised of three approximately 90\,mm long McKibben actuators. 
The amount of three individual actuators coincides with the minimum number of actuators necessary for a pulling mechanism to achieve planar circular motion. 
Force transfer is facilitated through strings that connect the ends of the actuators with the central aluminum tube of the actor.
It is possible to convert the coordinated alternating pressurization of the McKibben actuators into the planar (circular) motion of the tactor by redirecting the strings with the help of pulleys that are arranged in a triangular shape around the actor.

\begin{figure}[t]
    \centering
    \includegraphics{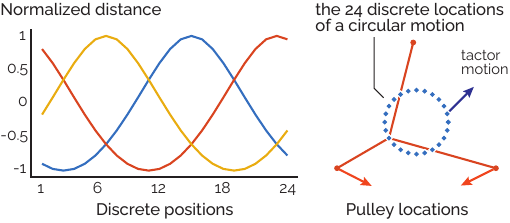}
    \caption{Sinusoidal phase-shifted normalized motor positions resulting from shortest distance calculations with 24 discrete positions. 
    The right plot shows an example of shortest distance calculation for one out of 24 discrete tactor positions. 
    Red dots show pulley locations, red lines show strings, and blue dots show the desired discretized tactor motion.}
    \label{fig:circ_mot}
\end{figure}

The circular massage movement is based on a precalculated list of motor positions.
The motion is modeled from shortest distances between the pulley locations and the discretized desired massaging motion.
The pulley wheels that guide the tactor-attached strings are located at the vertices of a triangle centered around the tactor’s axis.
By dividing the desired circular tactor motion into 24 discrete locations, we obtain the necessary respective string shortening or lengthening to achieve desired motion (Fig.~\ref{fig:circ_mot}).
The resulting motor commands lead to length changes of the McKibben actuators and thereby to the desired tensile forces and displacement of the massage head through the connected strings.

The designed feedback control loop aims to keep the tactor’s contact pressure at a constant setpoint value of 7.3\,kPa throughout the 0.1\,Hz circular motion as well as throughout the entire duration of the massage. 
This specific pressure value of 7.3\,kPa was determined empirically, demonstrating its efficacy in closely replicating the pressure exerted during a human massage
For this purpose, a custom pressure monitoring device feeds into a proportional controller which is implemented such that it outputs an updated positional term that is added to the previous motor command.
Moreover, it is known that silicone tubes can cause pressure losses in the overall system~\cite{pressure_drop}.
To prevent that we have added an additional parameter to the proportional term in the control model that accounts for pressure drop along the transmission path.
By incorporating this, we aim to compensate for the effects of pressure loss, thereby enhancing the precision and efficiency of the system's operation. 
The updated term is a proportionally scaled value of the error between the current tactor pressure and the setpoint pressure. 
In practice, this means that the syringe is compressed and pressure is increased (the tactor head is inflated) if the contact pressure is too low and vice versa.
Also, we limit the upper and lower absolute position output due to safety concerns.

\begin{figure}[t]
    \centering
    \includegraphics{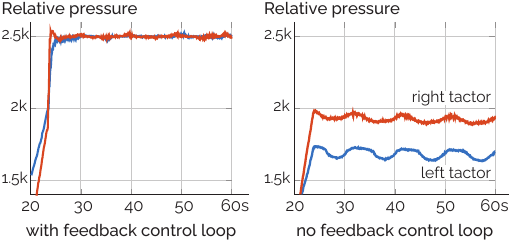}
    \caption{Comparison of tactor pressure measurements with and without \projname's feedback control loop.
    The system performed 3.5 circular massaging motions in the shown interval with a target contact pressure of 2.5\,kPa.
    When the feedback control loop is inactive, the tactor pressure decreases by 25\% compared to the target pressure while becoming more unstable.}
    \label{fig:controller}
\end{figure}

Fig.~\ref{fig:controller} compares the relative pressure applied by our robotic system to temple regions when we apply a proportional controller that is added to the previous motor command.
When a proportional controller is not applied to the robotic system (Fig.~\ref{fig:controller}b), the pressure value cannot reach the desired pressure, which is set to 2.5\,kPa for this experiment, while oscillating constantly.
However, when a feedback control loop is used to keep the tactor at constant pressure, the oscillations of relative pressure values diminish significantly while reaching the desired pressure.

\subsection{Participants}
For our experiments, we recruited nine participants with no known diseases; six males and three females with a mean age of 32 years and obtained written informed consent from all participants while the study adhered to the standard of the Declaration of Helsinki.

\subsection{Experimental Protocol and Data Collection} 
All of the study participants are treated equally and received the same massage intervention.
Before starting the massage participants were asked to sit down on a comfortable chair while the experimenter explained the procedure of the data collection with sensors to be used during the experiment.
After this preparation phase, data acquisition started with a 10-minute baseline period, followed by a 12-minute massage intervention and another 5-minute baseline period.
The participants were asked to sit quietly and refrain from talking during the whole procedure. 

\begin{figure}[b]
    \centering
    \includegraphics{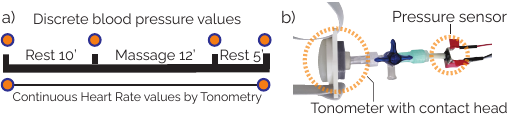}
    \caption{a) Timeline of the 27-minute study protocol with blood pressure (BP) and heart rate (HR) measurements and continuous tonometer measurement.
    b) Tonometer with contact head and loose mounting straps on the left and pressure sensor on the right.}
    \label{fig:exp_setup}
\end{figure}

During experiments, we record BP and HR with a Silvercrest HealthForYou blood pressure monitor and continuous pulse pressure waveform recordings with a self-developed tonometer placed on the Carotid artery, shown in Fig.~\ref{fig:exp_setup}b).
The custom pressure monitor device consists of a hollow silicone dome fixed to a mounting
platform and connected by tubing to an analog pressure sensor (ABPDANT060MGAA5, Honeywell) with a measurement range from 0 to 6\,kPa.
The silicone dome is positioned centrally over the carotid artery and transmits the arterial vessel deflections through pressure waves in the air inside the tubing to the pressure sensor.
We set the sampling frequency to 200\,Hz, which is high enough to extract HR~\cite{enough_fs}.
Overall the experiments took 27 minutes with a continuous tonometer and four discrete blood pressure measurements acquired as shown in Fig.~\ref{fig:exp_setup}.

\begin{figure}[t]
    \centering
    \includegraphics{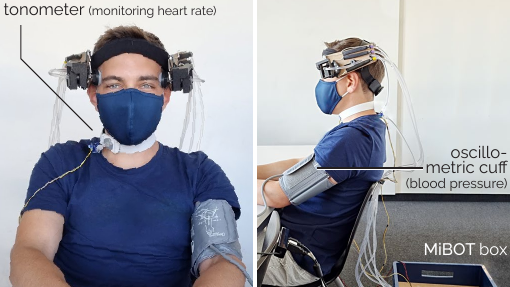}
    \caption{The experimental setup during \projname's evaluation included a tonometer on the carotid artery to measure heart rate as well as an oscillometric cuff to take blood pressure measurements.}
    \label{fig:study}
\end{figure}

Also, we compare \projname\ with a commercial massage device that uses air pressure and vibration to stimulate the scalp, mainly the forehead, and relax the muscles while having three intensity levels with two different temperature values.
At the beginning of the experiments, participants choose the massage modes, specifically the intensity and temperature levels according to their personal preferences.
We follow the same experimental protocol as in our designed massage robot to have a fair comparison.

\subsection{Data processing}
\label{sec:post_processing}
To extract HR from tonometry measurements, we first applied a third-order Butterworth bandpass filter (3--15\,Hz) to denoise the signal.
Then, we detected heartbeats in the filtered signals by finding peaks while restricting the minimum distance to 300\,ms, corresponding to a maximum HR of 180 beats-per-minute (bpm).
After finding these peaks, we smoothed the HR values and filtered out wrongly detected peaks due to motion artifacts such as transitions between massage sessions by averaging 10 adjacent beat-to-beat intervals.
We have not applied any processing to the obtained BP measurements as the output values from the device are processed internally.

\section{Results and Discussion}
\label{sec:Results}





A positive effect of successful massage therapy for headaches, especially prophylaxis of migraine attacks, can be indicated by a quantitatively measurable relaxation response caused by the massaging intervention~\cite{Diego_migraine}.
Since the massage increases the flow of venous blood and decreases the resistance of blood in arteries, which makes the heart-pumping action easier~\cite{WALASZEK2015396}. 
This could entail a decrease in heart rate and blood pressure values for both systole and diastole~\cite{decrease_in_BP}. 
To investigate \projname's effect on blood pressure, we analyze the measured systolic and diastolic BP values.

We averaged both systolic blood pressure (SBP) and diastolic blood pressure (DBP) for the first two measurements to get a baseline before the massage and did the same for the last two BP measurements to obtain values that correspond to the state after the massage.

\begin{figure}[t]
    \centering
    \includegraphics{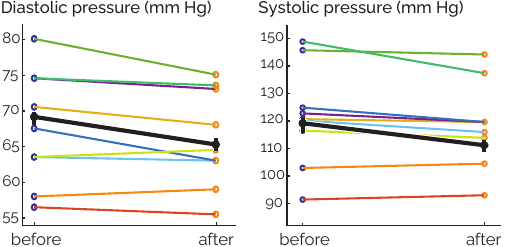}
    \caption{How participants' DBP (left) and SBP values (right) changed during the study.
    Colored lines represent participants, black shows the mean.}
    \label{fig:BP_results1}
\end{figure}

\begin{figure}[t]
    \centering
    \includegraphics{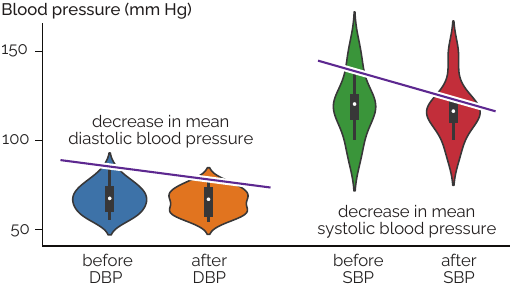}
    \caption{The statistical violin plots for blood pressure (DBP (left) and SBP (right)) given in direct comparison with before applying the massage and after. 
    The plot indicates the positive effect of massage on BP values for participants. 
    The violin plot includes the box plot (defined as Q1 and Q3 quartiles, and median) with a kernel density estimation over the points.}
    \label{fig:BP_violin}
\end{figure}

As shown in Fig.~\ref{fig:BP_results1}, the mean DBP and SBP values decreased after the massage was applied by our robot. 
While the mean value for DBP was 67.8\,mmHg before the massage, this value decreased to 66.1\,mmHg after the massage.
Similarly, SBP decreased from  119.8\,mmHg before the massage to 117.0\,mmHg after.
Although the mean BP shows a decreasing trend, the DBP or SBP values increased for three and two participants respectively.
However, the increase for those participants was smaller compared to the observed decrease for the other participants:

The maximum increase for the DBP and SBP were 1\,mmHg and 1.5\,mmHg, respectively.
Whereas, the maximum decrease was 5\,mmHg and 11\,mmHg for DBP and SBP respectively.
The overall distribution of blood pressure values before and after the massage with a kernel density estimation over the measured values and report is shown in the box plot in Fig.~\ref{fig:BP_violin} .
As shown in this figure, the decrease in DBP and SBP values is more pronounced at higher starting values.

The BP results, as shown in Fig.~\ref{fig:BP_results1} and~\ref{fig:BP_violin}, indicate that our designed massage robot has a positive effect on most of the participant's blood pressure, showing a similar relation as with human (manual) massage therapy.

\begin{figure}[b]
    \centering
    \includegraphics{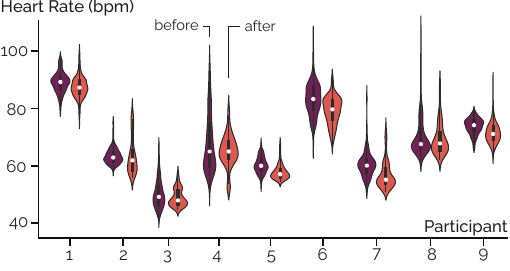}
    \caption{Violin plots showing the effect of \projname\ on participants' HR after applying the massage. 
    The HR's distribution mean decreased in six out of nine participants.
    Box plots inside the violins (Q1 and Q3 quartiles, and median) show a kernel density estimation over the points. 
    White circles show each participant's median before and after the massage.}
    \label{fig:HR_violin}
\end{figure}

Since a decrease in heart rate values is an indicator of a successful massage outcome, we analyzed the recorded tonometry values and obtained the HR in beats-per-minute (Section~\ref{sec:post_processing}).
HR distributions before and after the massage are shown as violin plots in Fig.~\ref{fig:HR_violin}.
To make the massage effect on HR comparable, we took sections with the same duration from the experiments for each participant. 
Therefore, before the massage,  HR was extracted from the first rest period, which corresponds to the first ten minutes of the experimental setup. 
To observe the effect of massage, we obtained the HR values from the last 5 minutes of the massage period with the final rest duration.

As shown in Fig.~\ref{fig:HR_violin}, seven out of nine participants experienced a decrease in HR after applying massage by the designed robotic system.
Overall, the mean decrease in HR values for all participants reaches 3.4\,bpm.
Although median values for Participant 3 increased after the massage, the maximum HR values decreased significantly.
Also, the increase in Participant 3 is 0.9\,bpm while the decrease in some participants reaches 5\,bpm (such as Participant 6) which is close to a 10\% decrease in HR on average.
These observed effects in HR and BP when \projname\ is applied for massage are similar to the human-administered massages by nurses~\cite{biosignal_embc}, which shows the effectiveness of \projname.
Also, when we compare the mean HR values for the baseline (10-minute rest period at the beginning of the data collection) and after the massage (5-minute rest period at the end of the data collection), we observed that there is a significant decrease (p$<$0.05) for six participants by \projname. 
This reaction is in accordance with the findings of previous works about massage therapy, where moderate human-administered massage decreased HRs and part of this effect was carried over to the post-session measurement as well~\cite{Diego_migraine}.


We also compare the effect of the massage applied by \projname\ with a commercial device that focuses on the temple and forehead regions while following the same experimental procedure.
Different from \projname, the commercial device offers two temperature options which are 40\textdegree\,C and 45\textdegree\,C.
Additionally, the device provides three distinct massage levels for the degree of vibrations, allowing participants to select their preferred temperature and intensity settings.
The changes in blood pressure and heart rate on average are given in Table~\ref{tab:comparison} while comparing with the effect of \projname.

\begin{table}[t]
\caption{\label{tab:comparison} Comparison of massage devices}
\centering
\renewcommand{\arraystretch}{1}
\begin{adjustbox}{width=0.8\columnwidth,center}
\begin{tabular}{@{}llll@{}}
\toprule
Variable/Device  & \projname\ & Commercial \\ \midrule
Heart rate  & -8.6\% $\pm$ 0.51 & +1.2\% $\pm$ 0.15  \\ 
Blood Pressure (SPB)  & -3.1\% $\pm$ 0.22 & -0.7\% $\pm$ 0.06  \\ 
Blood Pressure (DBP)  & -1.7\% $\pm$ 0.13 & -0.3\% $\pm$ 0.05  \\ 
\bottomrule
\end{tabular}
\end{adjustbox}
\end{table}

As shown in Table~\ref{tab:comparison}, the commercial massage device did not produce a relaxation effect;
no statistically significant difference between the means of baseline and after the massage was found for all temperature and intensity settings across subjects.
Furthermore, in contrast to the anticipated outcome, the commercial massage device actually resulted in an increased mean heart rate when we compared the baseline and after the massage.
Whereas, when wearing \projname, participants experienced an average reduction in heart rate of 8 to 10 percent, showing a significant performance improvement in providing relaxation.

While both devices concentrate on massaging the head area, these differences in findings can be explained if we look at the distinctions between \projname\ and the commercial device.
Our proposed massage device is different from the commercial device in several aspects. 
First, \projname\ uses tactor actuators that can create various tactile sensations with several angles for circular massage motion to the scalp.
These sensations are more natural and realistic than air pressure and vibration and can mimic the human touch. 
Second, our system has a silent actuation system that reduces the noise level of the massage. 
Commercial devices that use air pressure for massage generate a significant amount of noise when in operation, creating an irritating and distracting experience for the user, particularly since the device is positioned in close proximity to the head.
However, since we utilize a hydraulic transmission in our proposed massage system, the impact of noise has been significantly reduced.


Our promising results raise several questions on how to improve \projname's performance and other approaches that focus on the temple region of the head. 
To better personalize the massaging experience, the tactor's pressure and speed at which the tactor moves could be tailored to each person as suggested by some study participants.
In our final design, it takes about ten seconds to complete a full circle that covers the whole range of motion.
This could be sped up by increasing the McKibben actuators’ contraction ratios, which directly depend on the rate of syringe compression driven by the stepper motors. 
Our current system is limited to a relatively slow motor speed with an additional 3:1 gear ratio in order to produce the forces necessary for this process. 
Fluidic cylinders with less resistance and/or stronger motors would allow more variability in tactor speed.
Further studies, which take these variables into account, can be undertaken to design next-generation massage robots while increasing the user experience by personalizing the overall system.

Integrating \projname\ with unobtrusive sensors for real-time head-based cardiovascular monitoring~\cite{glabella} presents another opportunity to enhance its therapeutic effectiveness.
Dynamically adapting massage experiences in response to observed heart rate and blood pressure changes could help achieve optimal stress relief and relaxation.
By directly linking massage adjustments to physiological feedback, we open up possibilities for more targeted and efficient interventions.

Finally, the structural soundness of the compliant mechanism made from polylactic acid (PLA) was experimentally tested during use.
Torsional stability was achieved through the adequate height of the flexible members and plastic deformation caused by too much bending is prevented by the base plate’s motion limitation effect.
Nevertheless, aging effects could potentially critically weaken the part’s high-stress regions and thus fatigue behavior is something that should be investigated further for more robust systems.
\section{Conclusion}
\label{sec:Conclusion}

We have proposed \projname, a massage robot that modulates the wearer's cardiovascular responses through actuation of their temple regions with soft tactors to imitate human-like touch for the purpose of alleviating migraines and tension. 
The design of our robotic device addresses multiple challenges, including operation at near silent noise levels, which is a key requirement for administering effective and relaxing massage at the head.
\projname\ also modulates the massaging tactor’s normal pressure and adapts to complex head target topologies as well as the device's conformance to a wide range of different head shapes.
Our experimental evaluation with nine participants quantified the effects of a 12-minute massage intervention through \projname\ by measuring participants' blood pressure and heart rate levels.
Participants' measured physiological responses showed that the applied massage had a positive quantitative effect on participants via a decrease in heart rate and blood pressure values, indicating a massage-induced relaxation response.




\balance
\bibliographystyle{IEEEtran}
\bibliography{IEEEabrv,sample}

\begin{thebibliography}{10}
\providecommand{\url}[1]{#1}
\csname url@rmstyle\endcsname
\providecommand{\newblock}{\relax}
\providecommand{\bibinfo}[2]{#2}
\providecommand\BIBentrySTDinterwordspacing{\spaceskip=0pt\relax}
\providecommand\BIBentryALTinterwordstretchfactor{4}
\providecommand\BIBentryALTinterwordspacing{\spaceskip=\fontdimen2\font plus
\BIBentryALTinterwordstretchfactor\fontdimen3\font minus
  \fontdimen4\font\relax}
\providecommand\BIBforeignlanguage[2]{{%
\expandafter\ifx\csname l@#1\endcsname\relax
\typeout{** WARNING: IEEEtran.bst: No hyphenation pattern has been}%
\typeout{** loaded for the language `#1'. Using the pattern for}%
\typeout{** the default language instead.}%
\else
\language=\csname l@#1\endcsname
\fi
#2}}

\bibitem{stovner_global_2022}
L.~J. Stovner, K.~Hagen, M.~Linde, and T.~J. Steiner, ``The global prevalence
  of headache: an update, with analysis of the influences of methodological
  factors on prevalence estimates,'' \emph{The Journal of Headache and Pain},
  vol.~23, no.~1, p.~34, Apr. 2022.

\bibitem{ATLAS}
W.~H. Organization, ``Atlas of headache disorders and resources in the world
  2011,'' p. A collaborative project of World Health Organization and Lifting
  The Burden, 2011.

\bibitem{migraine_headache}
A.~Chaibi, P.~J. Tuchin, and M.~B. Russell, ``\BIBforeignlanguage{en}{Manual
  therapies for migraine: a systematic review},''
  \emph{\BIBforeignlanguage{en}{The Journal of Headache and Pain}}, vol.~12,
  no.~2, pp. 127--133, Apr. 2011.

\bibitem{EDVINSSON2012319}
L.~Edvinsson, C.~M. Villalón, and A.~MaassenVanDenBrink, ``Basic mechanisms of
  migraine and its acute treatment,'' \emph{Pharmacology \& Therapeutics}, vol.
  136, no.~3, pp. 319--333, 2012.

\bibitem{brain_int}
J.~Hoffmann, S.~M. Baca, and S.~Akerman, ``Neurovascular mechanisms of migraine
  and cluster headache,'' \emph{Journal of Cerebral Blood Flow \& Metabolism},
  vol.~39, no.~4, pp. 573--594, 2019, pMID: 28948863.

\bibitem{old_massage}
K.~Jones and W.~Du, ``Development of a massage robot for medical therapy,'' in
  \emph{Proceedings 2003 IEEE/ASME International Conference on Advanced
  Intelligent Mechatronics (AIM 2003)}, vol.~2, 2003, pp. 1096--1101 vol.2.

\bibitem{ICRA_peristaltic}
M.~Zhu, A.~Ferstera, S.~Dinulescu, N.~Kastor, M.~Linnander, E.~W. Hawkes, and
  Y.~Visell, ``A peristaltic soft, wearable robot for compression therapy and
  massage,'' \emph{IEEE Robotics and Automation Letters}, vol.~8, no.~8, pp.
  4665--4672, 2023.

\bibitem{ICRA_emg}
R.~C. Luo and C.~C. Chang, ``Electromyographic evaluation of therapeutic
  massage effect using multi-finger robot hand,'' in \emph{2011 IEEE
  International Conference on Robotics and Automation}, 2011, pp. 2431--2436.

\bibitem{massage_frontiers}
C.~Li, A.~Fahmy, S.~Li, and J.~Sienz, ``An enhanced robot massage system in
  smart homes using force sensing and a dynamic movement primitive,''
  \emph{Frontiers in Neurorobotics}, vol.~14, 2020.

\bibitem{ICRA_HRV_back}
Y.~Hasegawa, T.~Ootsuka, T.~Fukuda, F.~Arai, and M.~Kawaguchi, ``A relaxation
  system adapting to user's condition-identification of relationship between
  massage intensity and heart rate variability,'' in \emph{Proceedings 2001
  ICRA. IEEE International Conference on Robotics and Automation (Cat.
  No.01CH37164)}, vol.~4, 2001, pp. 3195--3200 vol.4.

\bibitem{IROS_EEG_massage}
R.~C. Luo, C.-W. Hsu, and S.-Y. Chen, ``Electroencephalogram signal analysis as
  basis for effective evaluation of robotic therapeutic massage,'' in
  \emph{2016 IEEE/RSJ International Conference on Intelligent Robots and
  Systems (IROS)}, 2016, pp. 2940--2945.

\bibitem{ICRA_lately}
R.~C. Luo, S.~Y. Chen, and K.~C. Yeh, ``Human body trajectory generation using
  point cloud data for robotics massage applications,'' in \emph{2014 IEEE
  International Conference on Robotics and Automation (ICRA)}, 2014, pp.
  5612--5617.

\bibitem{ICRA_contact}
Y.~Huang, P.~Souères, and J.~Li, ``Contact dynamics of massage compliant
  robotic arm and its coupled stability,'' in \emph{2014 IEEE International
  Conference on Robotics and Automation (ICRA)}, 2014, pp. 1499--1504.

\bibitem{hair_wash}
T.~Hirose, S.~Fujioka, O.~Mizuno, and T.~Nakamura, ``Development of
  hair-washing robot equipped with scrubbing fingers,'' in \emph{2012 IEEE
  International Conference on Robotics and Automation}, 2012, pp. 1970--1975.

\bibitem{Iros_new}
H.~Dong, Y.~Feng, C.~Qiu, Y.~Pan, M.~He, and I.-M. Chen, ``Enabling massage
  actions: An interactive parallel robot with compliant joints,'' in \emph{2022
  IEEE/RSJ International Conference on Intelligent Robots and Systems (IROS)},
  2022, pp. 4632--4637.

\bibitem{Dong_IROS}
------, ``Enabling massage actions: An interactive parallel robot with
  compliant joints,'' in \emph{2022 IEEE/RSJ International Conference on
  Intelligent Robots and Systems (IROS)}, 2022, pp. 4632--4637.

\bibitem{ICRA_soft}
C.~J. Payne, E.~G. Hevia, N.~Phipps, A.~Atalay, O.~Atalay, B.~R. Seo, D.~J.
  Mooney, and C.~J. Walsh, ``Force control of textile-based soft wearable
  robots for mechanotherapy,'' in \emph{2018 IEEE International Conference on
  Robotics and Automation (ICRA)}, 2018, pp. 5459--5465.

\bibitem{Kume}
M.~Kume, Y.~Morita, Y.~Yamauchi, H.~Aoki, M.~Yamada, and K.~Tsukamoto,
  ``Development of a mechanotherapy unit for examining the possibility of an
  intelligent massage robot,'' in \emph{Proceedings of IEEE/RSJ International
  Conference on Intelligent Robots and Systems. IROS '96}, vol.~1, 1996, pp.
  346--353 vol.1.

\bibitem{EMG_massage}
R.~C. Luo and C.~C. Chang, ``Electromyographic signal integrated robot hand
  control for massage therapy applications,'' in \emph{2010 IEEE/RSJ
  International Conference on Intelligent Robots and Systems (IROS)}, 2010, pp.
  3881--3886.

\bibitem{ECG_massage}
C.-C. Peng, T.-S. Hwang, C.-J. Lin, Y.-T. Wu, C.-Y. Chang, and J.-B. Huang,
  ``Development of intelligent massage manipulator and reconstruction of
  massage process path using image processing technique,'' in \emph{2010 IEEE
  Conference on Robotics, Automation and Mechatronics}, 2010, pp. 551--556.

\bibitem{IROS_Normal_Massage}
Y.~Huang, J.~Li, Q.~Huang, and C.~Liu, ``Design and control of anthropomorphic
  bit soft arms for tcm remedial massage,'' in \emph{2013 IEEE/RSJ
  International Conference on Intelligent Robots and Systems (IROS)}, 2013, pp.
  1960--1965.

\bibitem{ICRA_Arm}
M.~Khoramshahi, G.~Henriks, A.~Naef, S.~S.~M. Salehian, J.~Kim, and A.~Billard,
  ``Arm-hand motion-force coordination for physical interactions with non-flat
  surfaces using dynamical systems: Toward compliant robotic massage,'' in
  \emph{2020 IEEE International Conference on Robotics and Automation (ICRA)},
  2020, pp. 4724--4730.

\bibitem{ICRA_back}
R.~C. Luo, C.~P. Tsai, and K.~C. Hsieh, ``Robot assisted tapping control for
  therapeutical percussive massage applications,'' in \emph{2017 IEEE
  International Conference on Robotics and Automation (ICRA)}, 2017, pp.
  3606--3611.

\bibitem{IROS_OLD}
P.~Minyong, T.~Miyoshi, K.~Terashima, and H.~Kitagawa, ``Expert massage motion
  control by multi-fingered robot hand,'' in \emph{Proceedings 2003 IEEE/RSJ
  International Conference on Intelligent Robots and Systems (IROS 2003) (Cat.
  No.03CH37453)}, vol.~3, 2003, pp. 3035--3040 vol.3.

\bibitem{elsevier_bp_hr}
N.~L. Nelson, ``Massage therapy: understanding the mechanisms of action on
  blood pressure. a scoping review,'' \emph{Journal of the American Society of
  Hypertension}, vol.~9, no.~10, pp. 785--793, 2015.

\bibitem{magnetic_actuators}
D.~Q. Truong, T.~Q. Thanh, and K.~K. Ahn, ``\BIBforeignlanguage{en}{Development
  of a novel linear magnetic actuator with trajectory control based on an
  online tuning fuzzy {PID} controller},''
  \emph{\BIBforeignlanguage{en}{International Journal of Precision Engineering
  and Manufacturing}}, vol.~13, no.~8, pp. 1403--1411, Aug. 2012.

\bibitem{Noise_Level}
\BIBentryALTinterwordspacing
``\BIBforeignlanguage{en-us}{What {Noises} {Cause} {Hearing} {Loss}? {\textbar}
  {NCEH} {\textbar} {CDC}},'' Nov. 2022. [Online]. Available:
  \url{https://www.cdc.gov/nceh/hearing_loss/what_noises_cause_hearing_loss.html}
\BIBentrySTDinterwordspacing

\bibitem{Mckibben}
M.~Iwaki, Y.~Hasegawa, and Y.~Sankai, ``Study on wearable system for daily life
  support using mckibben pneumatic artificial muscle,'' in \emph{2010 IEEE/RSJ
  International Conference on Intelligent Robots and Systems (IROS)}, 2010, pp.
  3670--3675.

\bibitem{McKibben_Japan}
S.~Wakimoto, K.~Suzumori, and T.~Kanda, ``Development of intelligent mckibben
  actuator,'' in \emph{2005 IEEE/RSJ International Conference on Intelligent
  Robots and Systems (IROS)}, 2005, pp. 487--492.

\bibitem{pressure_drop}
C.~Asbach, H.~Kaminski, Y.~Lamboy, U.~Schneiderwind, M.~Fierz, and A.~M. Todea,
  ``Silicone sampling tubes can cause drastic artifacts in measurements with
  aerosol instrumentation based on unipolar diffusion charging,'' \emph{Aerosol
  Science and Technology}, vol.~50, no.~12, pp. 1375--1384, 2016.

\bibitem{enough_fs}
S.~Mahdiani, V.~Jeyhani, M.~Peltokangas, and A.~Vehkaoja, ``Is 50 hz high
  enough ecg sampling frequency for accurate hrv analysis?'' in \emph{2015 37th
  Annual International Conference of the IEEE Engineering in Medicine and
  Biology Society (EMBC)}, 2015, pp. 5948--5951.

\bibitem{Diego_migraine}
M.~Hernandez-reif, J.~Dieter, T.~Field, B.~Swerdlow, and M.~Diego, ``Migraine
  headaches are reduced by massage therapy,'' \emph{International Journal of
  Neuroscience}, vol.~96, no. 1-2, pp. 1--11, 1998.

\bibitem{WALASZEK2015396}
R.~Walaszek, ``Impact of classic massage on blood pressure in patients with
  clinically diagnosed hypertension,'' \emph{Journal of Traditional Chinese
  Medicine}, vol.~35, no.~4, pp. 396--401, 2015.

\bibitem{decrease_in_BP}
C.~M. Olney, ``The effect of therapeutic back massage in hypertensive persons:
  A preliminary study,'' \emph{Biological Research For Nursing}, vol.~7, no.~2,
  pp. 98--105, 2005, pMID: 16267371.

\bibitem{biosignal_embc}
T.~Ando, M.~Takeda, T.~Maruyama, Y.~Susuki, T.~Hirose, S.~Fujioka, O.~Mizuno,
  K.~Yamada, Y.~Ohno, and H.~Yukio, ``Biosignal-based relaxation evaluation of
  head-care robot,'' in \emph{2013 35th Annual International Conference of the
  IEEE Engineering in Medicine and Biology Society (EMBC)}, 2013, pp.
  6732--6735.

\bibitem{glabella}
C.~Holz and E.~J. Wang, ``Glabella: Continuously sensing blood pressure
  behavior using an unobtrusive wearable device,'' \emph{Proc. ACM Interact.
  Mob. Wearable Ubiquitous Technol.}, vol.~1, no.~3, sep 2017.

\end{thebibliography}

\end{document}